\documentclass[a4paper, 10pt, conference]{ieeeconf}      
\usepackage{FG2020}
\pdfoutput=1

\usepackage[normalem]{ulem}
\FGfinalcopy 

\IEEEoverridecommandlockouts                              
\usepackage{graphics} 
\usepackage{epsfig} 
\usepackage{mathptmx} 
\usepackage{times} 
\usepackage{amsmath} 
\usepackage{amssymb}  

\usepackage{booktabs}

\makeatletter
\let\NAT@parse\undefined
\makeatother

\usepackage[pagebackref=true,breaklinks=true]{hyperref}
\hypersetup{
	plainpages=false,
	colorlinks=true,              %
	linkcolor=blue,               %
	anchorcolor=blue,             %
	citecolor=blue,               %
	filecolor=blue,               %
	urlcolor=blue,                %
	pdfview=FitH,                 %
	pdfstartview=FitH,            %
	pdfpagelayout=SinglePage      %
}

\usepackage{multirow}

\usepackage{tikz,pgfplots}

\def\FGPaperID{142} 

\title{\LARGE \bf
	Can We Read Speech Beyond the Lips?\\Rethinking RoI Selection for Deep Visual Speech Recognition
}

\author{\parbox{15cm}{\centering
		{\large Yuanhang Zhang$^{1,2}$, Shuang Yang$^{1}$, Jingyun Xiao$^{1,2}$, Shiguang Shan$^{1,2}$, Xilin Chen$^{1,2}$}\\[2pt]
		{\normalsize
			$^1$ Key Laboratory of Intelligent Information Processing of Chinese Academy of Sciences (CAS), Institute of Computing Technology, CAS, Beijing 100190, China\\
			$^2$ University of Chinese Academy of Sciences, Beijing 100049, China}}
		\thanks{This work was done by Yuanhang Zhang during his internship at Institute of Computing Technology, Chinese Academy of Sciences.}
}

\begin{document}
	
	\ifFGfinal
	\thispagestyle{empty}
	\pagestyle{empty}
	\else
	\author{Anonymous FG 2020 submission\\ Paper ID \FGPaperID \\}
	\pagestyle{plain}
	\fi
	\maketitle
	
	\begin{abstract}
		Recent advances in deep learning have heightened interest among researchers in the field of visual speech recognition (VSR). Currently, most existing methods equate VSR with automatic \textit{lip reading}, which attempts to recognise speech by analysing lip motion. However, human experience and psychological studies suggest that we do not always fix our gaze at each other's lips during a face-to-face conversation, but rather scan the whole face repetitively. This inspires us to revisit a fundamental yet somehow overlooked problem: can VSR models benefit from reading extraoral facial regions, i.e. beyond the lips? In this paper, we perform a comprehensive study to evaluate the effects of different facial regions with state-of-the-art VSR models, including the mouth, the whole face, the upper face, and even the cheeks. Experiments are conducted on both word-level and sentence-level benchmarks with different characteristics. We find that despite the complex variations of the data, incorporating information from extraoral facial regions, even the upper face, consistently benefits VSR performance. Furthermore, we introduce a simple yet effective method based on Cutout to learn more discriminative features for face-based VSR, hoping to maximise the utility of information encoded in different facial regions. Our experiments show obvious improvements over existing state-of-the-art methods that use only the lip region as inputs, a result we believe would probably provide the VSR community with some new and exciting insights.
	\end{abstract}

	\section{Introduction}
	Visual speech recognition (VSR) is the task of recognising speech by analysing video sequences of people speaking. A robust VSR system has a variety of useful applications, such as silent speech interfaces~\cite{DBLP:conf/uist/0003YSLS18}, audio-visual speech recognition (AVSR) in noisy environments~\cite{afouras2018pami}, face liveness detection, and so on. Its performance has progressed significantly over the past few years, thanks to several successful deep learning architectures, such as convolutional neural networks (CNNs) and recurrent neural networks (RNNs). It also benefits from the emergence of large-scale, in-the-wild audiovisual datasets, from which deep neural networks can automatically learn strong representations that outperform previous hand-crafted features.
	
\begin{figure}
		\centering
		\includegraphics[width=\linewidth]{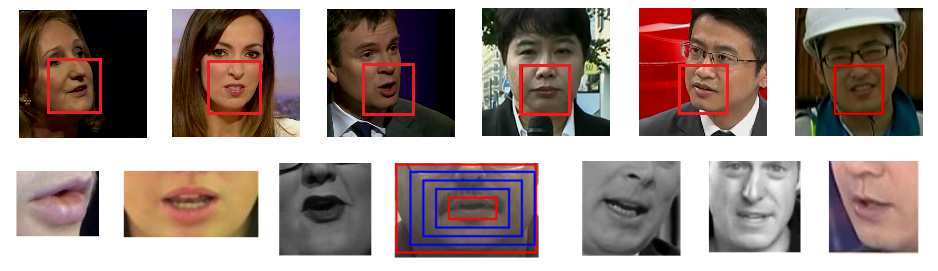}
		\caption{\textbf{Common practices for RoI selection in VSR.}
			To date, there is no clear consensus on a best practice, resulting in very different RoIs in different works (see Sec.~\ref{sec:related/vsr}). Top row: examples of frames from talking face videos. Bottom row: some examples of cropped RoIs in prior work, from left to right: \cite{DBLP:conf/fgr/AninaZZP15,DBLP:journals/corr/AssaelSWF16,DBLP:conf/accv/ChungZ16,DBLP:conf/avsp/KoumparoulisPMR17,DBLP:conf/interspeech/StafylakisT17,afouras2018pami,DBLP:conf/fgr/YangZFYWXLSC19}.}
		\label{fig:vsr_practice}
		\vspace*{-0.8cm}
	\end{figure}

	Traditionally, the term ``visual speech recognition" is used almost interchangeably with ``lip reading" within the VSR community, since it is usually believed that lip shapes and lip motion most likely contain almost all the information correlated with speech. The information in other facial regions is considered by default weak, and not helpful for VSR in practical use, due to the diversity of the speaker's pose and other variations in the facial region that are unrelated to speech production. Accordingly, as part of the dataset creation pipeline, almost all researchers crop regions-of-interest (RoIs) around the mouth after obtaining face bounding boxes and landmarks. By observing only the cropped RoI, the model is expected to focus on fine-grained discrimination of ``clean" motion signals within the most relevant lip region, and not be distracted by other parts of the face, whose utilities are less obvious. Some common practices for RoI selection in previous work are depicted in Fig.~\ref{fig:vsr_practice}.
	
	However, this convention of explicit mouth RoI cropping inevitably brings about many questions. Firstly, lip motion is not the only visual signal we can rely on to decode speech. Research on the advantage of performing speechreading with the whole face has a long history~\cite{sumby1954visual,benoit1996components}. In particular, movements of articulatory muscles such as the orbicularis oris (which is near the lips) lead to skin movements, often reflected by the cheeks being pushed and pulled, as well as changes in the visibility of the nasolabial folds. The now widely adopted term ``speechreading", used in place of ``lip reading" implies precisely the contribution of extraoral facial regions, such as the tongue, teeth, and cheeks, to the speech perception task. Evidence from psychology studies, and our experience in human communication also suggest that we in fact do not focus on the speaker's lip all the time throughout a conversation, even in very noisy environments. Instead, we scan different regions of the person's face periodically~\cite{vo2012eyes}.
	
	Secondly, if we do use a mouth RoI, there are many factors that need to be considered: how much of the face should the RoI cover? Will increased spatial context improve performance by supplying more information, or hinder performance by distracting the network with information unrelated to speech? Should we apply additional corrections to handle pose variation? Answers to these questions are not evident or straight, and there has been no consensus or a universal guideline for choosing RoIs until now. In fact, RoIs are often chosen based on intuition and the researchers' experience. The choices can be very different for different datasets and different methods, making transfer learning and cross-dataset evaluation difficult. Meanwhile, this specialised RoI selection step separates VSR from other face-related tasks, hindering further joint analysis. Using mouth inputs alone makes VSR an isolated problem, while including other facial regions opens up the possibility of various research in affective analysis and visual speech understanding. For example, joint modeling of non-verbal behaviour and verbal behaviour has been shown to be beneficial for learning adaptive word representations~\cite{DBLP:conf/aaai/WangSLLZM19}.
	
	Finally, data encountered in real-world applications may not have the same luxury that some academic datasets enjoy, where data is biased towards frontal or near-frontal views, and high-resolution mouth crops are easily obtainable. Models trained on such well-behaved, cropped data may not perform as well when presented with the various challenges in practice. This severely limits the potential usage scenarios of VSR systems.
	
	Motivated by these observations, we conduct a comparative study on input RoI selection, to (a) quantitatively estimate the contribution of different facial regions to the VSR task, and (b) determine whether state-of-the-art VSR networks, when presented with complex in-the-wild data, still benefit from additional clues within extraoral regions. Previous studies have explored different RoIs in the lower face~\cite{DBLP:conf/avsp/KoumparoulisPMR17,DBLP:conf/fgr/ShiraishiS15}, and demonstrated the importance of suitable RoI coverage and selection. However, these attempts are limited to the lower face, and relatively small datasets with narrow vocabularies. For this study, we approach the problem with state-of-the-art, deep VSR models trained on large-scale ``in-the-wild" VSR datasets, which depict many real-world variations, such as pose, lighting, scale, background clutter, makeup, expression, different speaking manners, etc. This is a much fairer reflection of real-world scenarios. Besides, we propose Cutout as an effective approach to encourage the model to utilise all facial regions. This allows researchers to sidestep the ambiguous choice of selecting appropriate RoIs, enhances the visual features, and increases the model's robustness to mild occlusion within the mouth or other facial regions.
	
	\section{Related Work}
	\subsection{Visual Speech Recognition}\label{sec:related/vsr}
	Visual speech recognition (VSR), commonly referred to as automatic lip reading, is a classical problem in computer vision and has received increased interest over recent years. The combination of deep learning approaches and large-scale audiovisual datasets has been highly successful, achieving remarkable word recognition rates and even surpassing human performance. The deep network approach has become increasingly common and mainstream~\cite{DBLP:conf/accv/ChungZ16,DBLP:journals/corr/AssaelSWF16,DBLP:conf/fgr/YangZFYWXLSC19}. These methods have retraced the success of the CNN-LSTM-DNN (CLDNN) paradigm~\cite{DBLP:conf/icassp/SainathVSS15} in the automated speech recognition (ASR) community. However, though many works have reported state-of-the-art results on multiple challenging datasets, few have investigated the influence of input RoI selection, which can be a nuisance due to variations in mouth shapes, face geometry, and pose, as well as the relatively unknown effect of spatial context and extraoral regions. Unlike face recognition, where face frontalisation is a recognised need for pose invariance, and has been thoroughly investigated as an important part of the pipeline, VSR researchers tend to specify RoIs based on their own experience, and in a dataset-specific manner.
	
	In some controlled-environment datasets, e.g. OuluVS~\cite{DBLP:journals/tmm/ZhaoBP09} and GRID~\cite{cooke2006audio}, the subjects remain relatively still while speaking. Lombard GRID~\cite{alghamdi2018corpus} uses a head-mounted camera which the speakers face directly at for the frontal view, essentially fully removing head motion. The Lip Reading in the Wild dataset~\cite{DBLP:conf/accv/ChungZ16} is of short duration with small or no face scale variation within clips, and loose registration is enforced by aligning nose centers. A fixed, mouth-centered (and sometimes affine-transformed) rectangular RoI is preferable on these datasets, because the faces are usually stable, frontal or near-frontal, and of uniform sizes. In contrast, OuluVS2~\cite{DBLP:conf/fgr/AninaZZP15} is a multi-view dataset, and RoIs are processed separately for each camera viewpoint. Chung and Zisserman~\cite{DBLP:conf/bmvc/ChungZ17} are the first to investigate large-scale deep lip reading in profile, and use an extended bounding box covering the whole face to account for significant pose variations. Yang et al.~\cite{DBLP:conf/fgr/YangZFYWXLSC19} determine the RoI size based on the distance between the nose and the mouth center and the width of the speaker's lips, which also effectively handles pose variations. 
	
	There have been some previous attempts to address the RoI selection problem explicitly. The most relevant work is \cite{DBLP:conf/avsp/KoumparoulisPMR17}, which perform experiments with rectangular lower face regions with different spatial context and resolution in a connected digits recognition task. \cite{DBLP:conf/fgr/ShiraishiS15} experiments with optical flow features in non-rectangular RoIs after removing head motion, and obtain results better than rectangular RoIs. However, these work are limited by the amount of data used and model capacity, and only investigate the utility of the lower face. We adopt face inputs, and mimic real-world scenarios by experimenting on large-scale, in-the-wild VSR benchmarks that are a magnitude larger than those used in the above two papers.
	
	\subsection{Human Perception Studies}
	Our intuition that we do not fix our gaze at the speaker's lip region when communicating with others is supported by a number of psychology studies on human gaze behaviour during visual and audio-visual speech perception~\cite{vatikiotis1998eye,vo2012eyes,lansing2003word}. It has been reported that human gaze patterns usually involve repetitive transitioning between the eyes and the mouth, even at high noise levels and when audio is absent~\cite{vatikiotis1998eye}. Interestingly, \cite{lansing2003word} suggests that in a visual-only scenario, speechreading accuracy was related to the difficulty of the presented sentences and individual proficiency, but not to the proportion of gaze time at a specific part of the face. The roles of the upper face, which is less intuitive for VSR, have also been studied~\cite{davis2006audio}. Studies in audiovisual speech perception show that head and eyebrow motion can help discriminate prosodic contrasts~\cite{cvejic2010prosody}, which may be helpful for the VSR task. Moreover, in tonal languages like Mandarin Chinese, movements in the neck, head, and mouth might be related to the lexical tones of syllables~\cite{chen2008seeing}. In the context of deep-learning based VSR, recently \cite{DBLP:conf/mmasia/ZhaoXS19} has shown that using video information from the cropped lip region together with \textit{pinyin} (Chinese syllables) information yields $0.85\%$ reduction in tone prediction error, supporting this hypothesis.
	
	\section{Exploring the Influence of RoIs on VSR}
	In this section, we first introduce the deep VSR architectures used in this work, and four manually selected RoIs (some of which include extraoral regions) to be experimented on. Next, we introduce Cutout as a simple strategy to enhance face-based VSR. Finally, we introduce a few visualisation methods we use to diagnose the resulting models.
	\subsection{Model Architecture}
	\textbf{3D-ResNet18.} The 3D-ResNet$18$ backbone which holds the current state-of-the-art on the LRW dataset~\cite{DBLP:conf/interspeech/StafylakisT17,DBLP:journals/cviu/StafylakisKT18} is used for all word-level experiments. It consists of a spatiotemporal convolution layer and a $18$-layer 2D residual network which gradually reduces spatial dimensionality, and yields a $512$-dimensional average-pooled vector for each frame. We use a $2$-layer bidirectional Gated Recurrent Unit (Bi-GRU) with $1024$ hidden units as the recurrent backend, and do not experiment with further regularisation such as Dropout and recurrent batch normalisation~\cite{DBLP:journals/cviu/StafylakisKT18}, as it deviates from the main objective of this paper.
	
	\textbf{LipNet.} For sentence-level VSR, we use the LipNet architecture, which achieves state-of-the-art performance on the GRID corpus~\cite{DBLP:journals/corr/AssaelSWF16}. With only three spatiotemporal convolutional layers in the frontend, this lightweight model should be less prone to overfitting on such a small-scale dataset. We also use temporal augmentation and Dropout as recommended by the authors.
	
	\subsection{Description of Manually Selected RoIs}\label{ssec:manual-roi}
	For fixed-size mouth crops, there are two dominant approaches: one uses fixed bounding box coordinates, and the other uses mouth-centered crops. To make this study comprehensive and self-complete, we experiment with both choices (although eventually we found no difference; see Table~\ref{table:lrw_baseline}).
	\begin{figure}
		\centering
		\includegraphics[width=0.95\linewidth]{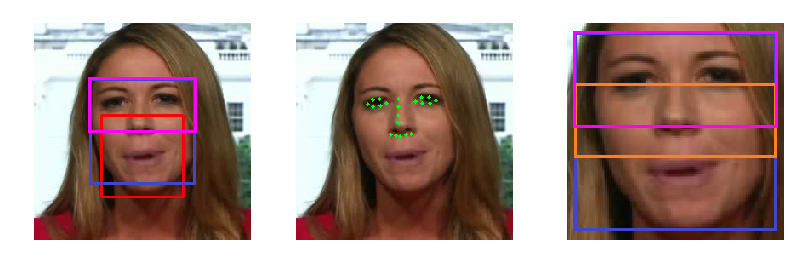}
		\vskip-0.5em
		{\qquad\small Original frame\hfill Landmark detection\hfill Aligned face\qquad}
		\vspace*{-0.2cm}
		\caption{\textbf{Illustration of the sub-face RoIs defined in this paper.}
			We train baselines on the whole face (blue), the upper face (purple), the cheeks (orange), and the mouth (red).}
		\label{fig:manual_roi_illustration}
		\vspace{-0.8cm}
	\end{figure}
	For face based models, we first crop and align the faces using detected eye and nose landmarks. This is because these parts undergo less non-rigid deformation, and this allows us to capture more significant motion in the lower face including the cheeks, the mouth and the jaw. At a low computational cost, this allows for more stable face tracks, and helps the network capture information in different spatial patches more consistently.
	However, it should be noted that face motion consists of rigid head movement (yaw, pitch, and roll) and non-rigid movement (e.g. frowning, raising eyebrows, and skin movement). Aligning the face retains yaw and pitch, but totally removes roll rotation. To this end, we also experiment with whole face and upper face cropped directly by fixed coordinates on LRW, since the faces are of uniform sizes and loosely aligned.
	The input sizes we choose are those commonly adopted in prior work, $112\times 112$ and $100\times 100$. We do not elaborate on the effects of spatial downsampling, since lower RoI resolution does not always yield poorer performance, as long as it remains above an acceptable level (e.g. $50\times 50$ pixels)~\cite{DBLP:conf/icip/BearHTL14,DBLP:conf/icip/DunganKH18,DBLP:conf/fgr/YangZFYWXLSC19}.
	Besides, in any case datasets collected in-the-wild already contain inherent scale and image quality variations.
	
	To crop the cheeks without revealing too much information in other regions, we crop a rectangular tile from the aligned face, which has no in-plane rotation. The vertical center of the RoI is set as the mean $y$ coordinate of the 18th, 27th, 57th, and 59th landmark after transformation.
	
	Finally, we do not experiment with the jaw and the lower neck. Although these parts are directly related to the vibrations of the vocal tract, these regions are not always visible, and such an RoI will be difficult to define.
	
	\subsection{Enhancing Face Based VSR with Cutout}
	Our motivation of using face as input is that it covers all other three sub-RoIs, which has the possibility of providing a strong ``holistic" representation. If the network is able to remain robust against the speech-irrelevant variations that human faces present, including pose, lighting, makeup, etc., we can benefit from additional information in other facial regions all at once.
	However, a network has only limited capacity. On a fixed budget, the goals of achieving invariance to nuisance factors and keeping stronger discrimination capabilities are inherently complimentary, and the recognition performance may suffer if either is not sufficiently realised.

	Indeed, our experiments will later show that the face models have already achieved performance comparable to lip based models. 
	However, when we inspect the convolutional responses of vanilla face-based models (see Fig.~\ref{fig:visualisation}) we find that the network tends to focus on the lip region and does not sufficiently utilise other parts of the face. Inspired by the observation that the vanilla face-based models able to achieve un-expected good performance, we try to enhance further the performance by asking the model to learn more discriminative features by utilising signals spread across the whole face.
	An intuitive idea is to create a strong patch-based ensemble, which has been shown feasible for facial expression recognition~\cite{DBLP:conf/icpr/LiZSC18}. However, for VSR it would be computationally expensive and impractical for deployment, since we are dealing with spatiotemporal data. Moreover, the redundancy between patches will burden optimisation.
	
	Therefore, we propose to apply Cutout~\cite{DBLP:journals/corr/abs-1708-04552}, a regularisation technique which has been popular with image CNNs. It augments the dataset with partially occluded versions of the samples in the dataset, and has already been successfully applied to image classification and object detection~\cite{DBLP:journals/corr/abs-1708-04896}. Our motivation is that this ``adversarial erasing" process should help the model pay more attention to less significant motion signals related to speech in extraoral regions.
	
	During training, a patch within the face region is randomly zeroed. Note that Cutout is applied to identical spatial positions across the video, since we expect the same facial region to be in roughly the same position after decent alignment. Over the course of the training process the model will encounter many masked versions of the same video, and eventually learn discriminative features for all parts of the face, which the model should be able to combine to its advantage during inference. Cutout fits seamlessly into the training process, and can be performed efficiently as a data augmentation step at no additional cost.
	
	\subsection{Interpreting Model Behaviour}
	To explore what our models are paying attention to throughout the video, we apply three visualisation techniques that can provide insights as to what the network focuses on for the task of visual speech recognition -- similar to the use of gaze trackers in human psychology experiments. 
	
	{\bf Feature maps.} Feature maps are filter responses to input images and outputs from previous layers, and can provide insight into intermediate representations learned by the model. Looking at responses from increasingly deeper layers can give us a sense of how the model combines low-level features into higher-level concepts that are useful for recognition.
	
	{\bf Saliency maps.} We use guided backpropagation saliency visualisation~\cite{DBLP:journals/corr/SpringenbergDBR14}, which has been previously utilised to interpret lip reading~\cite{DBLP:journals/corr/AssaelSWF16,DBLP:conf/accv/ChungZ16} and image classification models. Specifically, let $\mathbf{x}\in\mathbb{R}^{T\times H\times W\times C}$ be an input video, $V$ be the alphabet or vocabulary, and $\phi$ be the non-linear function that underlies the deep recognition network. For word-level classification, the network outputs a score for the $i$-th class in the vocabulary, $p(i\mid \mathbf{x}) = \phi(\mathbf{x})_i$, where $1\le i\le |V|$. We compute its gradient with respect to the input $\mathbf{x}$ using guided backpropagation. Likewise, for sentence-level VSR, we compute the likelihood of the greedily decoded output $\mathbf{y}\in (V\cup\{\text{\textvisiblespace}\})^*$, which is $\prod_t p(\mathbf{y}_t\mid \mathbf{x})$, and derive it against the input video sequence $\mathbf{x}$ to obtain the saliency maps. Eventually we obtain a tensor the same size as the input, which depicts spatial regions the model bases its prediction on for different timesteps.
	
	{\bf Spatiotemporal masking.} This approach, adapted from the patch masking method for visualising the receptive fields of 2D ConvNets~\cite{DBLP:conf/eccv/ZeilerF14} has been used to visualise important space-time video patches in audio-visual speech enhancement models~\cite{DBLP:journals/tog/EphratMLDWHFR18}. In our experiments, we mask each frame at the identical position for spatially aligned faces using a small $7\times 7$ patch in a sliding window fashion, and measure how overall accuracy is affected by computing performance drop, e.g. $\Delta_{\text{accuracy}}$. This process results in a heatmap depicting the contribution of different facial regions to recognition.
	\section{Experiments and Results}
	We train and evaluate on three VSR benchmarks, which cover tonal and atonal languages as well as in-the-wild and scripted speech: the Lip Reading in the Wild (LRW) dataset, the recent \textit{LRW}-1000 dataset, and the GRID audiovisual corpus. In this section, we first briefly introduce the three datasets we use, and present some implementation details. Next, we compare recognition performance using the four manually selected RoIs in Sec.~\ref{ssec:manual-roi}. We highlight the benefits of incorporating extraoral regions, in particular by using aligned entire faces. Finally, we present results of our best performing model which combine Cutout with face inputs, and make a few useful remarks.
	\subsection{Datasets}
	{\bf LRW.} LRW~\cite{DBLP:conf/accv/ChungZ16} is a challenging ``in-the-wild" English word-level lip reading dataset derived from BBC news collections, with $500$ classes and over $500,000$ instances, of which $25,000$ are reserved for testing.
	
	{\bf \textit{LRW}-1000.} \textit{LRW}-1000~\cite{DBLP:conf/fgr/YangZFYWXLSC19} is a $1000$-class, large-scale, naturally distributed Mandarin Chinese word-level lip reading dataset, also derived from TV broadcasts. With over $700,000$ word instances, it is even more challenging, with significant pose, scale, background clutter, word length, and inter-speaker variations. Note that the whole face was not provided with \textit{LRW}-1000 in the initial release. We will release face tracks for \textit{LRW}-1000\footnote{For now, we train and evaluate our models with the original word alignment and landmark annotations in \cite{DBLP:conf/fgr/YangZFYWXLSC19} for fair comparison. Note that the original paper used ResNet-$34$, while this work uses ResNet-$18$ which is currently more popular due to fewer parameters and better performance.}
	along with annotations that have undergone another round of manual cleaning as well as corresponding baselines in due course.
	
	{\bf GRID.} The GRID audiovisual corpus~\cite{cooke2006audio}, released in 2006, is a popular benchmark for sentence-level VSR. It consists of video recordings from $34$ speakers\footnote{One speaker's data is unavailable due to technical reasons.}, yielding $33,000$ utterances. All sentences follow a fixed grammar.
	\subsection{Implementation Details}
	{\bf Data preprocessing.} We detect faces and facial landmarks with the open-source SeetaFace2 toolkit~\cite{seetaface2019}, and align faces with similarity transformations using upper face landmarks~\cite{DBLP:conf/bmvc/ChungJZ17}, which are smoothed using a temporal Gaussian kernel of width $3$. For LRW and \textit{LRW}-1000, the faces are resized to $122\times 122$ and randomly cropped to $112\times 112$ during training. For GRID, the faces are resized to $100\times 100$.
	
	{\bf Training details.} All models are implemented with PyTorch and trained on $2$ to $3$ NVIDIA Titan X GPUs, each with $12$GB memory. We use the Adam optimizer with default hyperparameters. For word-level VSR, we use three-stage training described in \cite{DBLP:conf/interspeech/StafylakisT17}. Weight decay is set to $0.0001$, and learning rate is initialised to $0.0003$ for stage I / II and $0.001$ for stage III, decreasing on log scale from the $5$th or the $10$th epoch onwards, depending on whether Cutout is applied. We apply identical spatial augmentation to every frame in the form of random cropping and random horizontal flipping during training, and take a central crop for testing. Also, for \textit{LRW}-1000 training, we initialise front-end weights from the corresponding best model on LRW, following \cite{DBLP:conf/fgr/YangZFYWXLSC19}. For sentence-level VSR, we use a fixed learning rate of $0.0001$ and use no weight decay. We also apply random horizontal flipping, but no random cropping is used because of relatively accurate and stable landmarking results.
	
	{\bf Data partitioning, and evaluation metrics.} We use the train / validation / test partitioning provided with LRW and \textit{LRW}-1000. For GRID, we use $255$ random sentences from each speaker for evaluation, and the remainder for training, the same as previous state-of-the-art \cite{DBLP:journals/corr/AssaelSWF16,DBLP:conf/fgr/XuLCW18}. The evaluation metrics for word-level and sentence-level tasks are classification accuracy and Word Error Rate (WER), respectively, where WER is defined as
	\[\texttt{WER} = \frac{\text{\# of substitutions, deletions, and insertions}}{\text{length of ground truth transcript}}.\]
	Hence lower WERs are preferred. Note that the two metrics are equivalent if one views word recognition as a one-word sentence recognition task, in the sense that $\texttt{Acc} = 1-\texttt{WER}$.
	\subsection{Experiments on Manually Selected RoIs}
	Baseline results on the manually defined RoIs are shown in Table \ref{table:lrw_baseline}, Table \ref{table:lrw1000_baseline}, and \ref{table:grid_baseline}. We analyze the results step by step for the rest of this subsection.
	\begin{table}[]
		\centering
		\caption{Evaluation on the LRW dataset with different RoI choices. The second group uses directly cropped (upper) faces while the third applies face alignment.}
		\label{table:lrw_baseline}
		\vspace*{-0.25cm}
		\resizebox{\columnwidth}{!}{
			\begin{tabular}{@{}cccc@{}}
				\toprule
				\textbf{Region} & \textbf{Resolution} & \textbf{Accuracy} & \textbf{Description} \\ \midrule
				Mouth	&   $88\times 88$     & $83.30\%$  &  Fixed bounding box~\cite{DBLP:conf/icassp/PetridisSMCTP18,DBLP:journals/cviu/StafylakisKT18}  \\
				%
				Mouth	&   $88\times 88$     & $83.30\%$  &  Mouth-centered~\cite{DBLP:conf/accv/ChungZ16} \\ \midrule
				Face	&   $112\times 112$   & $83.46\%$    &  Nose-centered, $7/8$ original size \\
				Upper face & $112\times 112$ & $42.28\%$ & Resized upper half from above\\\midrule
				Face	&   $112\times 112$   & $83.10\%$    &  Aligned with eye \& nose landmarks~\cite{DBLP:conf/bmvc/ChungJZ17} \\
				Cheeks & $112\times 112$ & $62.49\%$ & Resized $40\times 112$ crop from above\\
				Upper face & $112\times 112$ & $48.33\%$ & Resized upper half\\
				Face (CBAM)	&   $112\times 112$   & $83.14\%$    &  \\
				\textbf{Face (Cutout)}	&   $112\times 112$   & $\bf 85.02\%$  &  \\
				\bottomrule
			\end{tabular}
		}
		\vspace*{-0.2cm}
	\end{table}
	\begin{table}[]
		\centering
		\caption{Evaluation on \textit{LRW}-1000 with different RoI choices.}
		\label{table:lrw1000_baseline}
		\vspace*{-0.25cm}
		\resizebox{\columnwidth}{!}{
			\begin{tabular}{@{}cccc@{}}
				\toprule
				\textbf{Region} & \textbf{Resolution} & \textbf{Accuracy} & \textbf{Description} \\ \midrule
				%
				Mouth	&   $88\times 88$  & $38.64\%$ &  Mouth-centered, no roll (as in~\cite{DBLP:conf/fgr/YangZFYWXLSC19})\\\midrule
				Face	&   $112\times 112$   & $41.71\%$   &  Aligned with eye \& nose landmarks \\
				Upper face	&   $112\times 112$   &  $15.84\%$  & Resized upper half\\
				Cheeks	&   $112\times 112$   & $32.50\%$    & Resized $40\times 112$ crop from above\\
				\textbf{Face (Cutout)}	&   $112\times 112$   & $\bf 45.24\%$    &  \\\midrule
				Upper face	&   $112\times 112$   &  $13.58\%$  & Front-end loaded from LRW and fixed \\
				\bottomrule
			\end{tabular}
		}
		\vspace*{-0.6cm}
	\end{table}
	\begin{table}[]
		\caption{Evaluation on the GRID corpus with different RoI choices. A $5$-gram, character-level language model is used during beam search. ``CER" stands for Character Error Rate. The lower the error rates, the better.}
		\label{table:grid_baseline}
		\centering
		\vspace*{-0.25cm}
		\resizebox{\columnwidth}{!}{
			\begin{tabular}{@{}ccccc@{}}
				\toprule
				\textbf{Region} & \textbf{Resolution} & \textbf{WER} & \textbf{CER} & \textbf{Description} \\ \midrule
				Mouth	&    $100\times 50$  &  $4.8\%$\cite{DBLP:journals/corr/AssaelSWF16} & $1.9\%$ &  Affine warped  \\
				Mouth	&    $100\times 50$  &  $4.7\%$ &  $1.9\%$  &  Above (reproduced)  \\
				Face	&   $100\times 100$   & $3.1\%$ & $1.3\%$  &  Aligned with eye \& nose landmarks \\
				Upper face	&   $100\times 50$   &   $14.4\%$  & $7.4\%$ & Upper half of above\\
				Cheeks & $100\times 50$ & $6.8\%$ & $3.1\%$ & Resized $36\times 100$ crop from above \\
				\textbf{Face (Cutout)}	&   $100\times 100$   & $\bf 2.9\%$ & $\bf 1.2\%$  &  \\\bottomrule
			\end{tabular}
		}
		\vspace{-0.7cm}
	\end{table}
	
	{\bf Effectiveness of including extraoral regions.} Experiment results clearly show that upper face and cheeks carry useful information, since recognition rates are far above chance. Counterintuitively, the upper face achieves nearly half the accuracy of face and mouth models. To ensure that the model has learned useful discriminative features instead of some unknown inherent bias within the dataset, we conduct an additional experiment, transferring from LRW to \textit{LRW}-1000 while keeping the front-end fixed. Intuitively, if the front-end has learned spurious clues that are irrelevant to the task itself, it should behave poorly on a dataset it has not been trained on. However, despite significant differences between the two datasets in terms of quality and language, classification accuracy is still far above chance after we fixed the front-end, with only $2.26\%$ absolute performance loss.
	
	Seeing how the upper face and cheeks convey useful information for VSR, by feeding the model with the entire face, we would expect it to benefit from the additional spatial context, which is indeed the case. Using the entire face instead of only the mouth region yields $1.6\%$ WER reduction on GRID, $3.07\%$ improvement on \textit{LRW}-1000, and $0.16\%$ improvement on LRW (when directly cropped faces are used). The slight performance regression on LRW with aligned faces will be discussed next.
	
	{\bf Making a case for face alignment.} For LRW, there are two sets of face-based experiments, one with face alignment, and the other using faces directly cropped by fixed coordinates. We observe a small performance degradation on LRW when we align faces to our canonical template ($0.2\%$ to $0.3\%$ drop relative to mouth and $7/8$ original resolution direct crops). Since recognition performance using the upper face actually benefits from alignment, it can be argued that the performance drop is most likely due to slightly lower mouth resolution (about $70\times 70$), and not removal of roll during the alignment process. This is not desirable, but acceptable, and there may be room for improvement if we adopt higher resolution inputs and improve landmarking quality. Therefore, we consider aligning the faces into stable face tracks to be beneficial, and use aligned faces for the remaining experiments. By any means, this is also necessary for structured facial region erasing with Cutout.
	
	{\bf Failure modes.} As an illustrative example, we further compare confusions made by the model under each crop setting in Table~\ref{table:top_pred_confusions_lrw} and Table \ref{table:preds_gt_grid} for the two English datasets. Overall, short words with less context and words that invoke weak extraoral motions perform worst. We observe that the words best predicted by the upper face are long words, such as ``Westminster" and ``Temperatures", and cheeks are good at making predictions for words that invoke significant extraoral motion. The face based model, which achieves comparable but slightly inferior performance compared to the mouth based model, fails to discriminate words with only subtle differences, such as ``benefits" and ``benefit". Recognising such words correctly requires analysis of tongue movement, which is hindered by the lowered overall resolution. However, on the GRID corpus, the face-based model seems to have identified idiosyncratic speaking styles, allowing it to make correct predictions for even short words and letters that are not easy to distinguish, and eventually obtain results better than cropped mouths.
	\begin{table}[]
		\caption{Top predicted words, worst predicted words and pairs exhibiting highest confusion in LRW under each crop setting.}
		\label{table:top_pred_confusions_lrw}
		\centering
		\vspace*{-0.2cm}
		\resizebox{\columnwidth}{!}{
			\resizebox{0.56\linewidth}{!}{
				\begin{tabular}{@{}cccc@{}}
					\toprule
					\multicolumn{4}{c}{\textbf{Accuracy / Network}} \\ \midrule
					\textbf{Mouth}  &  \textbf{Face}  &  \textbf{Cheeks}   &  \textbf{Upper Face}  \\ \midrule
					AGREEMENT (1) & ACCUSED (1) & AFTERNOON (0.98) & WESTMINSTER (0.96)\\\midrule
					ALLEGATIONS (1) & AGREEMENT (1) & WEEKEND (0.98) & TEMPERATURES (0.92)\\\midrule
					BEFORE (1) & BEFORE (1) & WELFARE (0.98) & AFTERNOON (0.88)\\\midrule
					PERHAPS (1) & CAMPAIGN (1) & WESTMINSTER (0.98) & SUNSHINE (0.88)\\\midrule
					PRIME (1) & FOLLOWING (1) & INFORMATION (0.96) & DESCRIBED (0.86)\\
					\bottomrule
				\end{tabular}
			}
			\hfill
			\resizebox{0.44\linewidth}{!}{
				\begin{tabular}{@{}cccc@{}}
					\toprule
					\multicolumn{4}{c}{\textbf{Accuracy / Network}} \\ \midrule
					\textbf{Mouth}  &  \textbf{Face}  &  \textbf{Cheeks}   &  \textbf{Upper Face}  \\ \midrule
					ASKED (0.58) & WORLD (0.6) & REALLY (0.32) & GREAT (0.18)\\\midrule
					BRITAIN (0.58) & ANSWER (0.58) & COULD (0.3) & OTHER (0.18)\\\midrule
					MATTER (0.58) & BECAUSE (0.58) & GETTING (0.3) & UNTIL (0.18)\\\midrule
					SPEND (0.58) & COURT (0.58) & MAKES (0.3) & WHICH (0.18)\\\midrule
					TAKEN (0.58) & PERSON (0.58) & MATTER (0.3) & BRING (0.16)\\
					\bottomrule
				\end{tabular}
			}			
		}
		\\[5pt]
		\resizebox{0.95\columnwidth}{!}{
			\begin{tabular}{@{}cccc@{}}
				\toprule
				\multicolumn{4}{c}{\textbf{Target / Estimated (\# Errors) / Network}}\\\midrule
				\textbf{Mouth} & \textbf{Face} & \textbf{Cheeks} & \textbf{Upper Face}\\\midrule
				SPEND / SPENT (11) & BENEFITS / BENEFIT (14) & CLAIMS / GAMES (11) & MILLION / BILLION (11) \\\midrule
				PRESS / PRICE (11) & PRESS / PRICE (12) & CHALLENGE / CHANGE (10) & BENEFITS / BENEFIT (10) \\\midrule
				WORST / WORDS (10) & LIVING / GIVING (12) & SYRIAN / SYRIA (10) & EVERYONE / EVERYBODY (10) \\\midrule
				PRICE / PRESS (10) & WORST / WORDS (11) & GROUND / AROUND (9) & TAKEN / SECOND (9) \\\midrule
				SERIES / SERIOUS (9) & THEIR / THERE (10) & INDUSTRY / HISTORY (9) & TERMS / TIMES (9) \\
				\bottomrule
			\end{tabular}
		}
		\vspace*{-0.2cm}
	\end{table}
	
	\begin{table}[]
		\caption{Examples of predictions on GRID under each crop setting. Prediction errors are highlighted in red.
			{\bf GT:} Ground Truth;
			{\bf UF:} Upper Face;
			{\bf C:} Cheeks;
			{\bf M:} Mouth;
			{\bf F:} (Aligned) Face.
		}
		\label{table:preds_gt_grid}
		\centering
		\vspace*{-0.3cm}
		\resizebox{0.4\columnwidth}{!}{
			\begin{tabular}{c|c}
				\toprule
				\textbf{GT} & lay white in u four now \\ \midrule
				\textbf{UF} &  lay white in u four now\\ \midrule
				\textbf{C} & lay white \textcolor{red}{at} \textcolor{red}{o} four now \\ \midrule
				\textbf{M} &  lay white \textcolor{red}{at} \textcolor{red}{o} four now \\ \midrule
				\textbf{F} & lay white in u four now \\ \bottomrule
			\end{tabular}
		}
		\qquad
		\resizebox{0.4\columnwidth}{!}{
			\begin{tabular}{c|c}
				\toprule
				\textbf{GT} & lay white in q five again \\ \midrule
				\textbf{UF} & \textcolor{red}{set} white \textcolor{red}{at} \textcolor{red}{t} five again \\ \midrule
				\textbf{C} & lay white \textcolor{red}{at} q five again \\ \midrule
				\textbf{M} & lay white \textcolor{red}{at} q five again \\ \midrule
				\textbf{F} & lay white in q five again \\ \bottomrule
			\end{tabular}
		}
		\vspace*{-0.7cm}
	\end{table}
	
	{\bf Superiority of using entire faces for cross-dataset transfer.} A byproduct of using the entire face rather than specifying sub-RoIs is that the visual front-end can be transferred across datasets easily with no explicit sub-RoI cropping after face detection and registration, which is already very routine in face recognition.
	Empirically, we found that transferring across face crops is much better than transferring across mouth crops from LRW to \textit{LRW}-1000. When both fine-tuned for $9$ epochs, mouth-based models suffer from source and target domain discrepancies, resulting in fluctuating validation loss and accuracy (best value $2.8482 / 38.40\%$), whereas the face-based model transferred stably, and converged to a better result (best value $2.7202 / 40.35\%$).
	\subsection{Experiments on Enhancing Face Inputs with Cutout}
\begin{figure}
		\centering
		\includegraphics[width=0.85\linewidth]{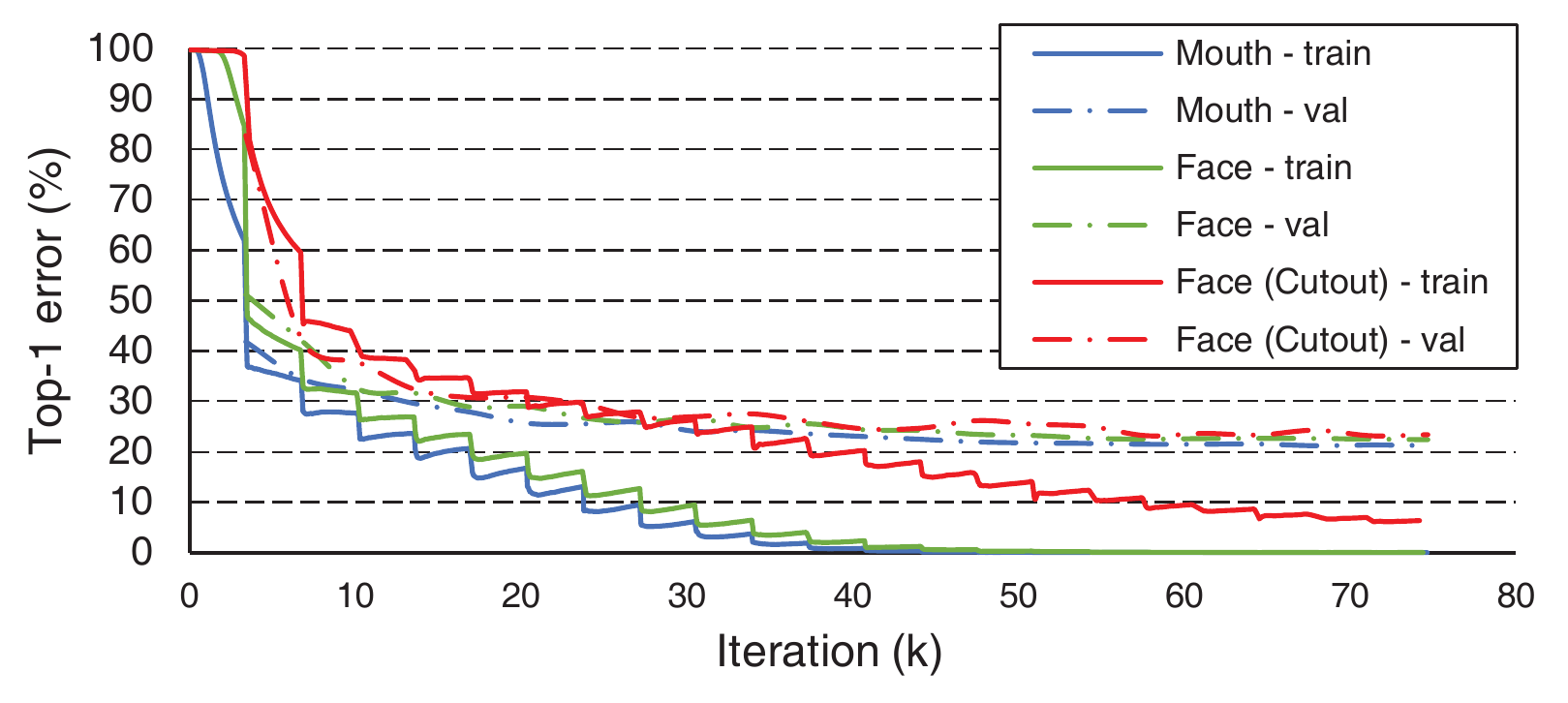}
		\vspace*{-0.4cm}
		\caption{\textbf{Accuracy curves on LRW as training progresses (only the first 75k steps of the temporal convolution backend stage is shown).}
			Compared to vanilla mouth and face based models which easily achieve nearly $100\%$ accuracy on the training set, Cutout (denoted by red curves) significantly reduces overfitting.}
		\label{fig:loss_curves}
		\vspace{-0.35cm}
	\end{figure}

	Results of combining Cutout with aligned faces can also be found in previous tables. This strategy is extremely powerful, enabling state-of-the-art performance on all three benchmarks. In particular, Cutout significantly reduces overfitting on LRW to the point where the model starts to underfit, as can be seen from the training curves in Fig.~\ref{fig:loss_curves}. This is radically different from models trained on faces or mouths without Cutout, where the network eventually reaches near $100\%$ accuracy on the training data, and proves that this harsh regularisation policy is indeed useful for VSR. Below we elaborate on a few interesting observations.
	
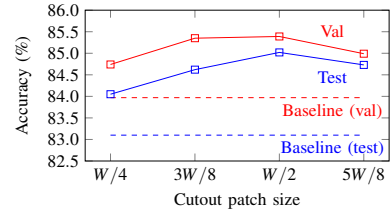
\begin{figure}
		\centering
		\resizebox{0.6\columnwidth}{!}{
			\begin{tikzpicture}
			\pgfplotsset{set layers}
			\begin{axis}[
			scale only axis,
			width=0.7*\linewidth,height=\linewidth*0.35,
			xlabel={Cutout patch size},
			ylabel={Accuracy (\%)},
			ymin=82.5, ymax=86,
			xtick={28,42,56,70},
			xticklabels={$W/4$,$3W/8$,$W/2$,$5W/8$},
			ytick={82.5, 83.0, 83.5, 84.0, 84.5, 85.0, 85.5, 86.0},
			legend columns=1,
			legend style={at={(0.5,-0.3)},anchor=north},
			y tick label style={
				/pgf/number format/.cd,
				fixed,
				fixed zerofill,
				precision=1,
				/tikz/.cd
			}
			]
			\addplot+[
			color=red,
			mark=square,
			]
			coordinates {
				(28,84.74)(42,85.35)(56,85.39)(70,84.99)
			} node[anchor=east,xshift=-0.5cm,yshift=0.2cm] (val) {};
			\node [above] at (val) {\textcolor{red}{Val}};
			\addplot+[
			color=blue,
			mark=square,
			]
			coordinates {
				(28,84.05)(42,84.62)(56,85.02)(70,84.73)
			} node[anchor=east,xshift=-0.5cm] (test) {};
			\node [below] at (test) {\textcolor{blue}{Test}};
			\addplot[
			color=red,dashed
			]
			coordinates {
				(28,83.97)(70,83.97)
			} node[anchor=east,xshift=-0.5cm] (baseval) {};
			\node [below] at (baseval) {\textcolor{red}{Baseline (val)}};
			\addplot[
			color=blue,dashed
			]
			coordinates {
				(28,83.1)(70,83.1)
			} node[anchor=east,xshift=-0.5cm] (basetest) {};
			\node [below] at (basetest) {\textcolor{blue}{Baseline (test)}};
			\end{axis}
			\end{tikzpicture}	
		}
		\vspace{-0.3cm}
		\caption{\textbf{Ablation results on LRW with different Cutout patch sizes.}
			We achieve best validation accuracy with patches half the size of the input.}
		\label{fig:cutout_ablation}
		\vspace{-0.8cm}
	\end{figure}

	{\bf Effect of the Cutout patch size.} The size of the masked out region is an important hyperparameter for Cutout. We experiment with four different sizes: $70\times 70$, $56\times 56$, $42\times 42$, and $28\times 28$, which are $5/8$, $1/2$, $3/8$, and $1/4$ the scale of the whole face, respectively. Experiments on LRW show that among those a $56\times 56$ patch is most effective (see Fig.~\ref{fig:cutout_ablation}). This is probably because it is approximately the average size of the mouth, and the possibility of the entire mouth being blocked allows for more efficient utilisation of extraoral regions. Since we adopt the same canonical template, we use $1/2$-size masks for all datasets (i.e. $50\times 50$ for GRID).
	
	\begin{figure*}[!ht]
		\centering
		\includegraphics[width=0.75\textwidth]{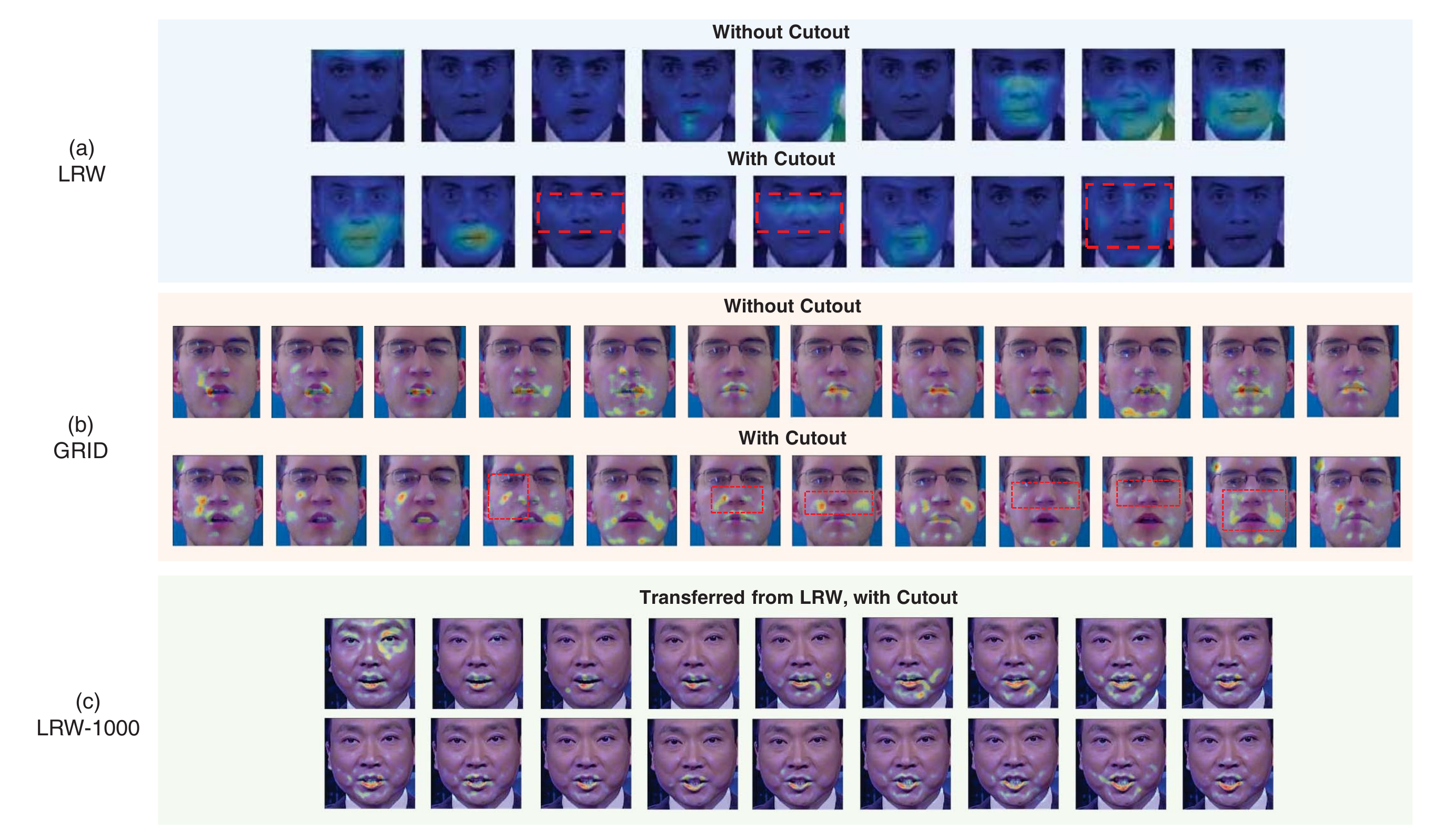}
		\vspace{-0.2cm}
		\caption{\textbf{Visualisations of model behaviour on LRW, \textit{LRW}-1000, and GRID with Cutout applied.}
			In particular, note how models trained with Cutout preserve fine-grained features, and yields clearly visible saliency around the cheeks. (a) ResNet layer 3 feature maps for ``hundred"; (b) Saliency maps for ``bin blue"; (c) Saliency maps for ``\textit{er ling yi qi}" (twenty-seventeen).}
		\label{fig:visualisation}
		\vspace{-0.3cm}
	\end{figure*}
	{\bf Visualising key facial regions.} Fig.~\ref{fig:visualisation} provide some saliency visualisations which show that models with Cutout can learn more discriminative features. Fig.~\ref{fig:visualisation}(a) is generated by extracting feature maps from the third ResNet layer, and performing max-pooling across channels, and Fig.~\ref{fig:visualisation}(b)(c) (GRID and \textit{LRW}-1000) by computing back-prop saliency. The derived maps are colored by intensity and overlaid onto the original frames. Each face thumbnail corresponds to a time step, and the two rows can be compared side-by-side to see the effects of introducing Cutout, especially regions highlighted with a dotted red box. For example, for LRW the convolutional responses are no longer confined to the lips, and for GRID there is stronger saliency in the cheeks, the eyebrows, and other facial muscles. The third row shows that after transferring from LRW to \textit{LRW}-1000, saliency in extraoral regions persist, which means that the learned facial features generalize well and are relatively robust.
	
\begin{figure}
		\centering
		\includegraphics[width=0.75\linewidth]{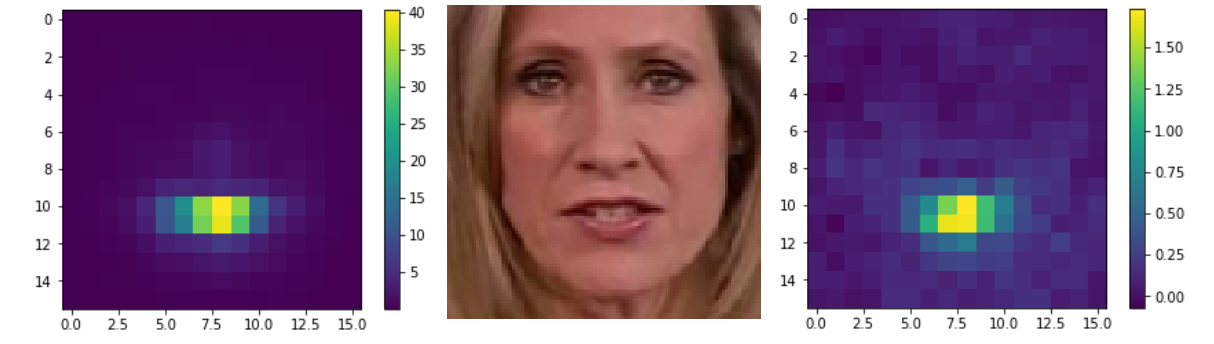}
		\vspace*{-0.25cm}
		\caption{\textbf{Important facial regions determined by spatiotemporal masking.}
			Left: without Cutout. Middle: an aligned face for reference. Right: with Cutout. Regions that result in more accuracy drop when occluded are colored brighter, and more crucial for model prediction.}
		\label{fig:face_distribution}
		\vspace{-0.7cm}
	\end{figure}

	We also identify regions that are important to the network's predictions by spatiotemporal masking on the entire test set of LRW, and results are depicted in Fig.~\ref{fig:face_distribution}. It can be seen that for both models, the area that leads to the most significant performance degradation when masked is still the lip region, which agrees with common intuition. In addition, the model trained with Cutout is also affected by occlusion in extraoral regions such as the cheeks and the upper face, showing that the model has learned strong visual features that encode these weak signals to complement lip motion. More importantly, while the plain model suffers up to $40\%$ performance drop even when only a $7\times 7$ patch is occluded, the drop remains below $2\%$ for the model trained with Cutout. This observation again strongly supports the usefulness of extraoral regions.
	
	{\bf Comparison with attention-based RoI selection.} The attention mechanism, inspired by human vision, has been widely used for implicit RoI selection and fine-grained image recognition. Here we compare Cutout with a Convolutional Block Attention Module (CBAM)~\cite{DBLP:conf/eccv/WooPLK18} augmented baseline on LRW, where we plug CBAM into ResNet blocks. Here, we use $5\times 5$ instead of $7\times 7$ kernels for the spatial attention modules to accommodate to the smaller feature maps. Although CBAM is a powerful spatial-channel attention mechanism which achieved remarkable performance on ImageNet classification and robust remote heart rate estimation~\cite{DBLP:conf/fgr/NiuZHDDSC19}, results show that the attention-augmented model is only marginally better than the baseline on LRW. We believe this is because the subtle movements in extraoral regions are too weak to be captured with the attention mechanism, and the model is still biased towards lip motion.
	
	{\bf Performance across pose.} We are also interested in how well the Cutout-augmented model can handle pose variations. We turn to \textit{LRW}-1000, where the data is divided into different difficulty levels according to yaw rotation. From Table \ref{table:perf_pose}, we can see that the model trained with Cutout outperforms the no-augmentation face-based baseline by about $2\%$ on both the easy (yaw $\ge 20^\circ$) and the medium ($\ge 40^\circ$) subset, but degrades slightly on the hard subset ($\ge 60^\circ$). We believe this is mainly because effective areas are smaller in the hard setting, where there are more frames in profile view. The effective regions are more likely to be erased or occluded when there is significant yaw rotation.
	
	%
	\begin{table}
		\caption{Performance w.r.t Pose on \textit{LRW}-1000}
		\label{table:perf_pose}
		\renewcommand{\arraystretch}{0.45}
		\vspace*{-0.2cm}
		\centering
		\begin{tabular}{ccccc}
			\toprule
			\textbf{Methods} & \textbf{Easy} & \textbf{Medium} & \textbf{Hard} & \textbf{All} \\ \midrule
			Mouth (ResNet34) \cite{DBLP:conf/fgr/YangZFYWXLSC19} & $24.89\%$ & $20.76\%$ & $15.9\%$ & $38.19\%$\\
			\midrule
			Mouth & $25.95\%$ & $21.55\%$ & $18.36\%$ & $38.64\%$\\
			\midrule
			Face & 28.87\% & 28.45\% & $\mathbf{27.21\%}$ & 41.71\%\\  
			\midrule
			\textbf{Face (Cutout)} & $\mathbf{31.74\%}$ & $\mathbf{30.04\%}$ & 26.89\% & $\mathbf{45.24\%}$ \\ 
			\bottomrule           
		\end{tabular}
		\vspace*{-0.6cm}
	\end{table}
	
	\section{Conclusion}
	In this paper, we have investigated a previously overlooked problem in VSR: the use of extraoral information. We demonstrate that extraoral regions in the face, such as the upper face and the cheeks, can also be included to boost performance. We show that using simple Cutout augmentation with aligned face inputs can yield stronger features, and vastly improve recognition performance by forcing the model to learn the less obvious extraoral cues from data. Beyond VSR, our findings also have clear implications for other speech-related vision tasks, such as realistic talking face generation, face spoofing detection and audio-visual speech enhancement. Next steps include extending the method to sentence-level VSR where more contextual clues are available, increasing input resolution, and eliminating the need for explicit face alignment.
	
	\ifFGfinal
	\section{Acknowledgments}
	We would like to thank Chenhao Wang and Mingshuang Luo's extensive help with data processing. This work is supported in part by the National Key R\&D Program of China (No. 2017YFA0700804), and Natural Science Foundation of China (No. 61702486, 61876171).
	\fi
	
	{\small
		\bibliographystyle{ieee}
		\bibliography{refs}
	}
	
\end{document}